\documentclass[11pt,a4paper]{article}
\usepackage[hyperref]{emnlp2020}
\usepackage{times}
\usepackage{latexsym}

\usepackage{microtype}
\usepackage{booktabs} 
\usepackage{url}
\usepackage{float}
\usepackage{graphicx}
\usepackage{multirow}
\usepackage{multicol}
\usepackage{balance}
\usepackage{amsfonts}
\usepackage{amsmath}
\usepackage{xcolor}
\newcommand*\circled[1]{\kern-2.5em%
  \put(0,4){\color{white}\circle*{18}}\put(0,4){\circle{10}}%
  \put(-3,0){\color{black}\bfseries#1}~~}
\usepackage{tikz} 
\newcommand*\circledd[1]{\tikz[baseline=(char.base)]{
            \node[shape=circle,draw,inner sep=.8pt] (char) {\bfseries#1};}}
\usepackage{enumitem}
\DeclareMathOperator*{\argmax}{argmax} 
\aclfinalcopy 

\aclfinalcopy 
\def\aclpaperid{} 

\newcommand{\modelname}{\texttt{SelfORE}}
\newcommand{\bert}{Contextualized Relation Encoder}
\newcommand{\clustering}{Adaptive Clustering}
\newcommand{\classification}{Relation Classification}
\title{{\modelname}: Self-supervised Relational Feature Learning for Open Relation Extraction}

\author{Xuming Hu$^1$, Chenwei Zhang$^{2\dagger}$, Yusong Xu$^1$, Lijie Wen$^{1\dagger}$, Philip S. Yu$^{1,3}$\\
  $^1$Tsinghua University,  $^2$Amazon, $^3$University of Illinois at Chicago\\
  $^1$\texttt{\{hxm19,xys20\}@mails.tsinghua.edu.cn}\\
  $^2$\texttt{cwzhang@amazon.com}
  $^1$\texttt{wenlj@tsinghua.edu.cn}  \\
  $^3$\texttt{psyu@uic.edu}\\
  }
 
\date{}

\begin{document}
\maketitle
\begin{abstract}
Open relation extraction is the task of extracting open-domain relation facts from natural language sentences. Existing works either utilize heuristics or distant-supervised annotations to train a supervised classifier over pre-defined relations, or adopt unsupervised methods with additional assumptions that have less discriminative power. In this work, we propose a self-supervised framework named {\modelname}, which exploits weak, self-supervised signals by leveraging large pretrained language model for adaptive clustering on contextualized relational features, and bootstraps the self-supervised signals by improving contextualized features in relation classification.
Experimental results on three datasets show the effectiveness and robustness of {\modelname} on open-domain Relation Extraction when comparing with competitive baselines. Source code is available\footnote{\url{https://github.com/THU-BPM/SelfORE}\\\phantom{00} $^\dagger$Corresponding Authors.}.
\end{abstract}

\section{Introduction}
With huge amounts of information people generate, Relation Extraction (RE) aims to extract triplets of the form (subject, relation, object) from sentences, discovering the semantic relation that holds between two entities mentioned in the text. For example, given a sentence \textit{Derek Bell was born in Belfast}, we can extract a relation \textsc{born in} between two entities \textit{Derek Bell} and \textit{Belfast}. 
The extracted triplets from the sentence are used in various 
down-stream applications like web search, question answering, and natural language understanding.  

Existing RE methods work well on pre-defined relations that have already appeared either in human-annotated datasets or knowledge bases. While in practice, human annotation can be labor-intensive to obtain and hard to scale up to a large number of relations. Lots of efforts are made to alleviate the human annotation efforts in Relation Extraction.
Distant Supervision \cite{mintz2009distant} is a widely-used method to train a supervised relation extraction model with less annotation as it only requires a small amount of annotated triplets as the supervision. However, distant supervised methods usually make strong assumptions on entity co-occurrence without sufficient contexts, which leads to noises and sparse matching results. More importantly, it works on a set of pre-defined relations, which prevents its applicability on open-domain text corpora.

Open Relation Extraction (OpenRE) aims at inferring and extracting triplets where the target relations cannot be specified in advance. Besides approaches that first identify relational phrases from open-domain corpora using heuristics or external labels via distant supervision and then recognize entity pairs \cite{yates2007textrunner,fader2011identifying}, clustering-based unsupervised representation learning models get lots of attentions recently due to their ability to recognize triplets from meaningful semantic features with minimized or even no human annotation. 
\citet{yao2011structured} regards OpenRE as a totally unsupervised task and uses clustering method to extract triplets with new relation types. However, it cannot effectively discard irrelevant information and select meaningful relations. \citet{simon2019unsupervised} trains expressive relation extraction models in an unsupervised setting.
But it still requires that the exact number of relation types in the open-domain corpus is known in advance.

To further alleviate the human annotation efforts while obtaining high-quality supervision for open relation extraction, in this paper, we propose a self-supervised learning framework which obtains supervision from the data itself and learns to improve the supervision quality by learning better feature presentations in an iterative fashion.
\begin{figure}[htbp!]
    \centering
    \includegraphics[width=\linewidth]{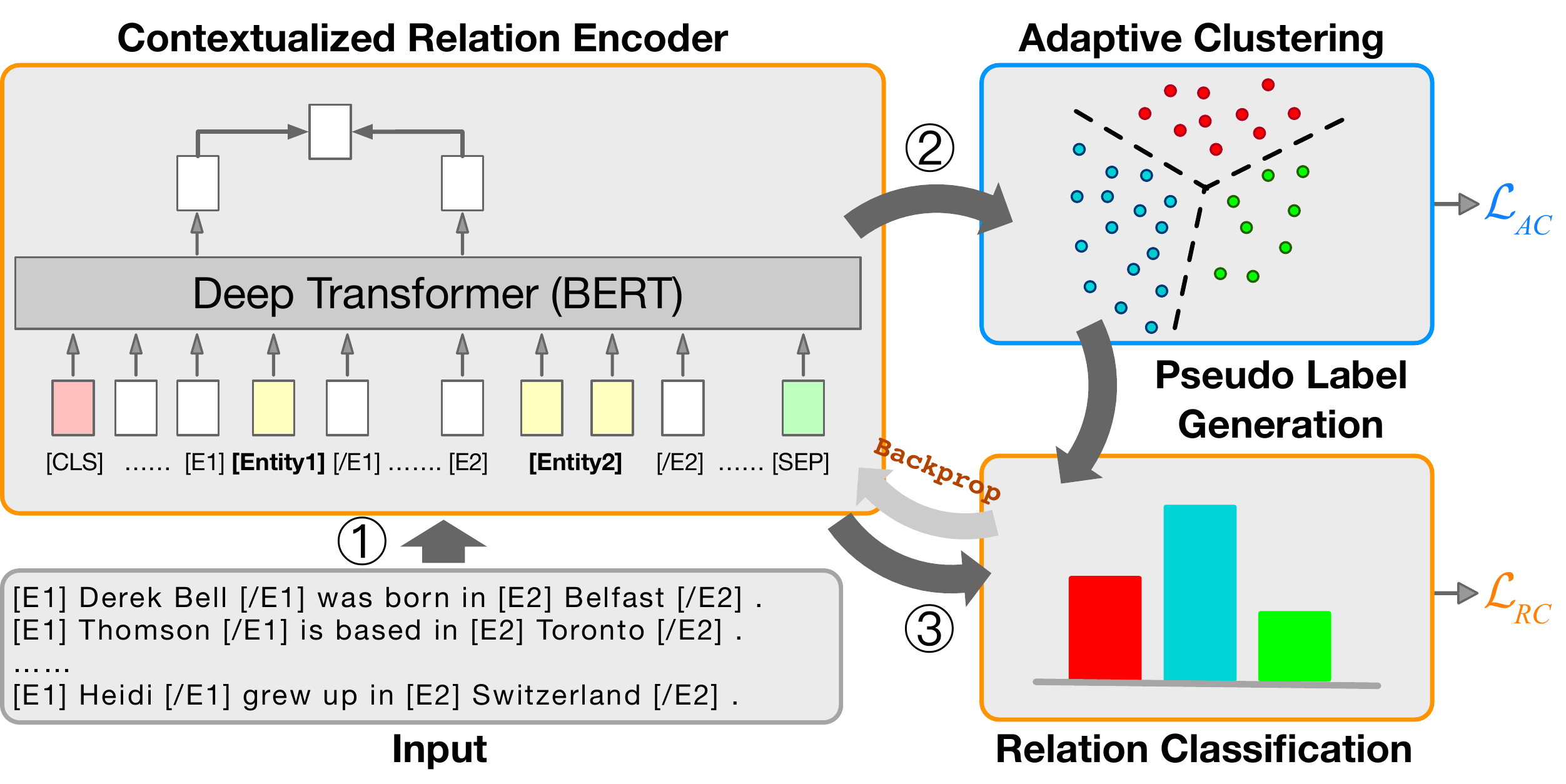}
    \caption{Open Relation Extraction via Self-supervised Learning.}
    \label{fig:overview}
\end{figure}
The proposed framework has three modules, {\bert}, {\clustering}, and {\classification}. As shown in Figure \ref{fig:overview}, the {\bert} leverages pretrained BERT model to encode entity pair representations based on the context in which they are mentioned. 
To recognize and facilitate proximity of relevant entity pairs in the relational semantic space, the {\clustering} module effectively clusters the contextualized entity pair representations generated by {\bert} and generates pseudo-labels as the self-supervision.
The {\classification} module takes the cluster labels as pseudo-labels to train a relation classification module. The loss of {\classification} on self-supervised pseudo labels helps improve contextualized entity pairs features in {\bert}, which further improves the pseudo label quality in {\clustering} in an iterative fashion.

To summarize, the main contributions of this work are as follows:
\begin{itemize}
    \item We developed a novel self-supervised learning framework {\modelname} for relation extraction from open-domain corpus where no relational human annotation is available.
    \item We demonstrated how to leverage pretrained language models to learn and refine contextualized entity pair representations via self-supervised training schema.
    \item We showed that the self-supervised model outperforms strong baselines, and is robust when no prior information is available on target relations.
\end{itemize}
\section{Proposed Model}
The proposed model {\modelname} consists of three modules: {\bert}, {\clustering}, and {\classification}. As illustrated in Figure \ref{fig:overview}, the {\bert} takes sentences as the input, where named entities are recognized and marked in the sentence. {\bert} leverages the pretrained BERT \cite{devlin2018bert} model to output contextualized entity pair representation. The {\clustering} takes the contextualized entity pair representation as the input, aiming to perform clustering that determines the relational cluster an entity pair belongs to. Unlike traditional clustering methods which assign hard cluster labels to each entity pair and are sensitive to the number of clusters, {\clustering} performs soft-assignment which encourages high confidence assignments and is insensitive to the number of clusters.
The pseudo labels based on the clustering results are considered as the self-supervised prior knowledge, which further guides the {\classification} and features learning in {\bert}.
 
Before introducing details of each module, we briefly summarize the overall learning schema:
\begin{enumerate}[label=\protect\circled{\arabic*}]
    \item Obtain contextualized entity pair representations based on entities mentioned in sentences using {\bert}.
    \item Apply {\clustering} based on updated entity pair representations in \circledd{1} to generate
    pseudo labels for all relational entity pairs.
    \item Use pseudo labels as the supervision to train and update both {\bert} and {\classification}. Repeat \circledd{2}.
\end{enumerate}

\subsection{\bert} 
The contextualized relation encoder aims to extract contextualized relational representations between two given entities in a sentence. In this work, we assume named entities in the text have been recognized ahead of time and we only focus on binary relations which involve two entities.

The type of relation between a pair of entities can be reflected by their contexts. Also, the nuances of expression in contexts also contribute to the relational representation of entity pairs. Therefore, we leverage pretrained deep bi-directional transformers networks \citep{devlin2018bert} to effectively encode entity pairs, along with their context information.

For a sentence ${X = [x_{1}, .., x_{T}]}$ where two entities $E1$ and $E2$ are mentioned, we follow the labeling schema adopted in \citet{soares2019matching} and augment ${X}$ with four reserved tokens to mark the beginning and the end of each entity mentioned in the sentence. We introduce the ${[E1_{start}]}$, ${[E1_{end}]}$, ${[E2_{start}]}$, ${[E2_{end}]}$ and inject them to $X$:
\begin{equation}
\begin{aligned}
{X}=&\big[x_{1},...,[E1_{start}],x_{i},...,x_{j-1},[E1_{end}],\\
\qquad...,&[E2_{start}],x_{k},...,x_{l-1},[E2_{end}],...,x_{T}\big]
\end{aligned}
\end{equation}
as the input token sequence for {\bert}. 

The contextualized relation encoder is denoted as $f_{\theta}(X,E1,E2)$.
To get the relation representation of two entities $E1$ and $E2$, instead of using the output of $[CLS]$ token from BERT which summarizes the sentence-level semantics, we use the outputs corresponding to ${[E1_{start}]}$ , ${[E2_{start}]}$ positions as the contextualized entity representation and concatenate them to derive a fixed-length relation representation $\mathbf{h}\in\mathbb{R}^{2\cdot{h_{R}}}$:
\begin{equation}
\mathbf{h} = [\mathbf{h}_{[E1_{start}]}, \mathbf{h}_{[E2_{start}]}].
\end{equation}

\subsection{\clustering} 
After we obtained $H=\{\mathbf{h}_1, \mathbf{h}_2, ..., \mathbf{h}_N\}$ from $N$ contextualized entity pair representations using {\bert}, {\clustering} aims to cluster entity pair representations into $K$ semantically-meaningful clusters. {\clustering} gives each entity pair a cluster label, which serves as the pseudo label for later stages.

Comparing with the traditional clustering method which gives hard label assignment for each entity pair (e.g. ${k}$-means), the {\clustering} adopts a soft-assignment, adaptive clustering schema. The adaptive clustering encourages high-confidence assignments and is insensitive to the number of clusters. More specifically, {\clustering} consists of two parts: (1) a non-liner mapping ${g_\phi}$ to convert the entity pair representation $\mathbf{h}\in\mathbb{R}^{h_{R}}$ to a latent representation $\mathbf{z}\in\mathbb{R}^{h_{AC}}$, (2) learning a set of ${K}$ cluster centroids ${\{\mathbf{\mu}_{k}\in \mathbb{R}^{h_{AC}}\}_{k=1}^{K}}$, and a soft-assignment of all $N$ entity pairs to $K$ cluster centroids.

For the first part, we simply adopt a set of fully connected layers as the non-linear mapping. Instead of initializing parameters randomly and training the mapping from scratch, the initial parameters are adopted from an encoder of an autoencoder model \cite{vincent2010stacked}. 
We pretrain an autoencoder model separately, which takes $\mathbf{h}$ as the input and minimizes the reconstruction loss over all $N$ samples:
\begin{align}
\widetilde{\mathbf{h}}=&Dropout(\mathbf{h}) \\
\mathbf{z}=&\text{g}(\mathbf{W}_{\phi}\widetilde{\mathbf{h}}+\mathbf{b}_{\phi})\\
\widetilde{\mathbf{z}}=&Dropout(\mathbf{z})\\
\hat{\mathbf{h}}=&\text{d}(\mathbf{W}_{\sigma} \widetilde{\mathbf{z}}+\mathbf{b}_{\sigma}).
\end{align}

For the second part, the module learns to optimize ${g_\phi}$'s parameters and assign each sample to a cluster with high confidence.
We first perform standard ${k}$-means clustering in the feature space $\mathbb{R}^{h_{AC}}$ to obtain $K$ initial centroids ${\{\mathbf{\mu} _{k}\in \mathbb{R}^{h_{AC}}\}_{k=1}^{K}}$. 
Inspired by \citet{xie2016unsupervised},
we use the Student's ${t}$-distribution as a kernel to measure the similarity between embedded point ${\mathbf{z}_{n}}$ and each centroid ${\mathbf{\mu}_{k}}$:
\begin{equation}
q_{nk}=\frac{(1+||\mathbf{z}_{n}-\mathbf{\mu}_{k}||^{2}/\alpha )^{-\frac{\alpha +1}{2}}}{\sum _{k'}(1+||\mathbf{z}_{n}-\mathbf{\mu}_{k'}||^{2}/\alpha )^{-\frac{\alpha +1}{2}}},
\end{equation}
where ${\alpha}$ represents the freedom of the Student's ${t}$-distribution and ${q_{nk}}$ can be regarded as the probability of assigning sample ${n}$ to cluster ${k}$ as the soft assignment. We set ${\alpha=1}$ for all experiments.

We normalize each cluster by frequency as an auxiliary target distribution in Equation \ref{eq::normalize_freq} and iteratively refine clusters by learning from high-confidence assignments with the help of an auxiliary distribution:
\begin{equation}\label{eq::normalize_freq}
p_{nk}=\frac{q_{nk}^{2}/f_{k}}{\sum_{k'}q_{nk'}^{2}/f_{k'}},
\end{equation}
where ${f_{k}={\sum} _{n}q_{nk}}$ is the soft cluster frequency.

With the auxiliary distribution, we define KL divergence loss between the soft assignments ${q_{n}}$ and the auxiliary distribution ${p_{n}}$ as follows to train the {\clustering} module:
\begin{equation}
\mathcal{L}_{AC}=KL(P||Q)=\sum _{n}\sum _{k}p_{nk}log\frac{p_{nk}}{q_{nk}}.
\end{equation}

We use gradient descent based optimizer to minimize $\mathcal{L}_{AC}$. Note that only the parameters for $g_{\phi}$ will be updated ---parameters in the {\bert} ($f_{\theta}$) are not effected when minimizing $\mathcal{L}_{AC}$. We assign the pseudo label $s_n$ for the $n$-th entity pair by taking the label associated with the largest probability:
\begin{equation}
    s_n = \argmax_{k \in \mathcal{K}} p_{nk}.
\end{equation}
To alleviate the negative impact from choosing unideal initial centroids, {\clustering} re-selects a set of $K$ initial centroids randomly if $\mathcal{L}_{AC}$ does not decrease after the first epoch.

In summary, comparing with traditional clustering methods such as $k$-means, {\clustering} adopts an iterative, soft-assignment learning process which encourages high-confidence assignments and uses high-confidence assignments to improve low-confidence ones. {\clustering} possesses the following advantages: 1) It improves clustering purity and benefits low-confidence assignment for an overall better relational clustering performance. 2) It prevents large relational clusters from distorting the hidden feature space. (3) It neither requires the actual number of target relations in advance (although it is good to have the target relations as the prior knowledge), nor the distribution of relations.

\subsection{\classification}
{\clustering} generates cluster labels $S = \{s_1, s_2, ..., s_N\}$ for all entity pairs as pseudo labels. With these pseudo labels as self-supervised signals derived from the corpora themselves, {\classification} module aims to use pseudo labels to guide the relational feature learning in {\bert} as well as relation classifier learning in {\classification}.

Similar to supervised classifiers which learn to predict golden labels, the {\classification} module learns to predict the pseudo labels generated by {\clustering}:
\begin{equation}
    \mathbf{l}_n = c_{\tau}(f_{\theta}(X_n, E1, E2)),
\end{equation}
where $c_{\tau}$ denotes the relation classification module parameterized by $\tau$ and $\mathbf{l}_n$ is a probability distribution over $K$ pseudo labels for the $n$-th sample.
In order to find the best-performing parameters $\theta$ for {\bert} and $\tau$ for {\classification}, we optimize the following classification loss:
\begin{equation}
\mathcal{L}_{RC} = \underset{\theta,\tau}{min}\frac{1}{N}\sum_{n=1}^{N}\mathit{loss}(\mathbf{l}_n,\operatorname{one\_hot}(s_{n})),
\end{equation}
where $loss$ is the cross entropy loss function and  $\operatorname{one\_hot}(s_n)$ returns a one-hot vector indicating the pseudo label assignments. 

\subsection{The Bootstrapping Self-Supervision Loop}\label{sec:rel_name}
After optimizing $\mathcal{L}_{RC}$, we repeat {\clustering} and {\classification} in an iterative fashion, shown as \circledd{2},\circledd{3} in Figure \ref{fig:overview}.
Overall, the {\clustering} exploits weak, self-supervised signals from data and {\classification} bootstraps the discriminative power of the {\bert} by improving contextualized relational features for Relation Classification.
Note that for {\clustering}, although it does not update {\bert}, it always utilizes the updated $\theta$ to get the most up-to-date entity pair feature representations $\mathbf{h}$ for clustering. Hence it generates stronger self-supervision as the loop goes on, by providing pseudo labels with higher quality for the {\classification} module.

We stop the clustering and classification loop when current pseudo labels have less than 10\% difference with the former epoch. To get the surface-form relation name for each cluster, if there is one, we get words between $[E1]$ and $[E2]$ and calculate the most frequent n-gram as the surface form. For quantitative evaluation, we assign the majority ground truth label within each cluster as the predict relation label for each relation cluster.

\section{Experiments}
We conduct extensive experiments on real-world datasets to show the effectiveness of our self-supervised learning rationale on relation extraction, and give a detailed analysis to show its advantages.
\subsection{Datasets}
Three datasets are used to evaluate our model: NYT+FB, T-REx SPO, and T-REx DS. 
NYT+FB dataset aligns sentences from the New York Times corpus \cite{sandhaus2008new} with Freebase \cite{bollacker2008freebase} triplets. It has been widely used in previous RE works \cite{yao2011structured,marcheggiani2016discrete,simon2019unsupervised}.
We follow the setting in \citet{simon2019unsupervised} and filter out sentences with non-binary relations. We get 41,000 labeled sentences containing 262 target relations from 2 million sentences. 20\% of these sentences will be used as validation datasets for hyperparameter tuning and 80\% will be used for model training.

Both T-REx SPO and T-REx DS datasets come from T-REx \cite{elsahar2018t} which is generated by aligning Wikipedia corpus with Wikidata \cite{vrandevcic2012wikidata}. We filter triplets and keep sentences where both entities appear in the same sentence --- a sentence will appear multiple times if it contains multiple binary relations associated with different entity pairs. We built two datasets T-REx SPO and T-REx DS depending on whether the dataset has surface-form relations or not. For example, the relation \textit{give birth to} could be conveyed by surface-forms like \textit{born in}, \textit{date of birth}, etc. T-REx SPO contains 615 relations and 763,000 sentences, where all sentences contain triplets having the surface form relation in the sentence. T-REx DS is generated where the surface-form of relation is not necessarily contained in the sentence. T-REx DS contains 1189 relations and nearly 12 million sentences. The dataset still contains some misalignment, but should nevertheless be easier for models to extract the correct semantic relation. 20\% of these sentences will be used as the validation dataset and 80\% will be used for model training.

\subsection{Baseline and Evaluation metrics}
We use standard unsupervised evaluation metrics for comparisons with other three baseline algorithms \citet{yao2011structured,marcheggiani2016discrete,simon2019unsupervised} where no human annotation is available for Relation Extraction from the open-domain data. For all models, we assume the number of target relations is known to the model in advance. We set the number of clusters to the number of ground-truth categories and evaluate performance with ${B}^{3}$, V-measure and ARI. 

Additionally, we evaluate the performance of our proposed model in a practical, yet more challenging setting: we assume the size of target relations is not known. A much larger cluster size $\hat K$ such as 1000 is adopted. When ${\hat K}\gg{K}$, we use unsupervised approaches such as ${k}$-means to further merge $\hat K$ clusters into $K$ clusters (the size of ground-truth categories) for a fair evaluation.

For baselines, rel-LDA is a generative model proposed by \citet{yao2011structured}. We consider two variations of rel-LDA which only differ in the number of features they considered. rel-LDA uses the 3 simplest features and rel-LDA-full is trained with a total number of 8 features listed in \citet{marcheggiani2016discrete}. UIE \cite{simon2019unsupervised} is the state-of-the-art method that trains a discriminative relation extraction model on unlabeled datasets by forcing the model to predict each relation with confidence and encourage all relations to be predicted on average. Two base model architectures (UIE-March and UIE-PCNN) are considered. To make it fair comparison, we further introduce UIE-BERT, which is trained with losses introduced in \citet{simon2019unsupervised} but we replace the PCNN classifier + GloVe embedding with our BERT-based Relation Encoder and Classification module.

To convert pseudo labels indicating the clustering assignment to relation labels for evaluation purposes, we follow the setting in the previous work \cite{simon2019unsupervised} and assign the majority of ground truth relation labels in each cluster to all samples in that cluster as the prediction label. 
For evaluation metrics, we use ${\text{B}^{3}}$ precision and recall to measure the correct rate of putting each sentence in its cluster or clustering all samples into a single class. More specifically, ${\text{B}^{3}}$ ${F_{1}}$ is the harmonic mean of precision and recall:
\[
\text{B}{^{3}}~\text{Prec.} =\underset{X,Y}{\mathbb{E}} P(g(X)=g(Y)|c(X)=c(Y))
\]
\[
\text{B}{^{3}}~\text{Rec.} =\underset{X,Y}{\mathbb{E}} P(c(X)=c(Y)|g(X)=g(Y)).
\]

We use V-measures \cite{rosenberg2007v} to calculate homogeneity and completeness, which is analogous to ${\text{B}^{3}}$ precision and recall, but with the conditional entropy:
\[
\text{Homogeneity} =1-H(c(X)|g(X))/H(c(X))
\]
\[
\text{Completeness} =1-H(g(X)|c(X))/H(g(X))
\]
where these two metrics penalize small impurities in a relatively ``pure'' cluster more harshly than in less pure ones. We also report F1, which is the harmonic mean of Homogeneity and Completeness.

Adjusted Rand Index \cite{hubert1985comparing} measures the degree of agreement between two data distributions. The range of ARI is [-1,1], the larger the value, the more consistent the clustering result is with the real situation. 

\begin{table*}[thbp!]
\centering
  \resizebox{.84\linewidth}{!}{%
\begin{tabular}{c|l|ccc|ccc|c}
\multirow{2}{*}{Dataset} & \multirow{2}{*}{Model}       & \multicolumn{3}{c|}{$B^{3}$}                       & \multicolumn{3}{c|}{V-measure}                                                           & \multirow{2}{*}{ARI}               \\ \cline{3-8}
                        &&F1            & Prec.         & Rec.          & F1            & Hom.          & Comp.         &                                    \\ \hline
\multirow{8}{*}{NYT+FB}  & {rel-LDA\cite{yao2011structured}}     & 29.1          & 24.8          & 35.2          & \multicolumn{1}{c}{30.0}          &     \multicolumn{1}{c}{26.1}          & 35.1          & \multicolumn{1}{c}{13.3}          \\ \cline{2-9} 
                         & {rel-LDA-full\cite{yao2011structured}}    & 36.9          & 30.4          & 47.0          & 37.4         & 31.9          & 45.1          & \multicolumn{1}{c}{24.2}          \\ \cline{2-9}
                         & March\cite{marcheggiani2016discrete}            & 35.2          & 23.8          & 67.1 & 27.0         & 18.6          & 49.6          & \multicolumn{1}{c}{18.7}          \\ \cline{2-9} 
                         & UIE-March\cite{simon2019unsupervised}              & 37.5          & 31.1          & 47.4          & 38.7          & 32.6          & 47.8          & \multicolumn{1}{c}{27.6}          \\ \cline{2-9} 
                         & UIE-PCNN\cite{simon2019unsupervised}   & 39.4          & 32.2          & 50.7          & 38.3          & 32.2          & 47.2          & 33.8          \\ \cline{2-9} 
                         & UIE-BERT& 41.5          & 34.6          & 51.8          & 39.9          & 33.9          & 48.5          & \multicolumn{1}{c}{35.1}          \\ \cline{2-9} 
                        & {{\modelname} w/o Classification} & 30.7 & 28.2 & 33.8        & 23.7 & 21.9 & 25.6          & \multicolumn{1}{c}{20.0} \\ \cline{2-9}
                         & {{\modelname} w/o {\clustering}} & 46.2 & 45.1 & 47.4          & 44.1 & 43.2 & 45.0         & \multicolumn{1}{c}{37.6} \\ \cline{2-9}
                        & {\modelname} & \textbf{49.1} & 47.3 & 51.1          & \textbf{46.6} & 45.7 & 47.6         & \multicolumn{1}{c}{\textbf{40.3}} \\ \hline \hline
                          
\multirow{8}{*}{T-REx SPO}  & {rel-LDA\cite{yao2011structured}}     & 11.9          & 10.2          & 14.1          & 5.9          & 4.9          & 7.4          & \multicolumn{1}{c}{3.9}          \\ \cline{2-9} 
                         & {rel-LDA-full\cite{yao2011structured}}    & 18.5          & 14.3          & 26.1          & 19.4         & 16.1         & 24.5          & \multicolumn{1}{c}{8.6}          \\ \cline{2-9} 
                         & March\cite{marcheggiani2016discrete}          & 24.8          & 20.6          & 31.3 & 23.6          & 19.1         & 30.6          & \multicolumn{1}{c}{12.6}          \\ \cline{2-9} 
                         & UIE-March\cite{simon2019unsupervised} & 29.5          & 22.7          & 42.0          & 34.8          & 28.4         & 45.1          & \multicolumn{1}{c}{20.3}          \\ \cline{2-9} 
                         & UIE-PCNN\cite{simon2019unsupervised}    & 36.3          & 28.4          & 50.3          & 41.1         & 33.7         & 53.6         & \multicolumn{1}{c}{21.3}          \\ \cline{2-9} 
                         & UIE-BERT  & 38.1          & 30.7          & 50.3          & 39.1          & 37.6          & 40.8          & \multicolumn{1}{c}{23.5}          \\ \cline{2-9} 
                        & {{\modelname} w/o Classification} & 32.7 & 28.3 & 38.6         & 25.3 & 23.1 & 28.0          & \multicolumn{1}{c}{22.5} \\ \cline{2-9}
                         & {{\modelname} w/o {\clustering}} & 34.5 & 31.2 & 38.5         & 29.2 & 27.4& 31.2          & \multicolumn{1}{c}{28.3} \\ \cline{2-9}
                        & {\modelname} & \textbf{41.0} & 39.4 & 42.8        & \textbf{41.4} & 40.3 & 42.5        & \multicolumn{1}{c}{\textbf{33.7}} \\ \hline \hline
\multirow{8}{*}{T-REx DS}  & {rel-LDA\cite{yao2011structured}}     & 9.7          & 6.8          & 17.0          & 8.3          & 6.6         & 11.4         & \multicolumn{1}{c}{2.2}          \\ \cline{2-9} 
                         & {rel-LDA-full\cite{yao2011structured}}    & 12.7          & 8.3          & 26.6          & 17.0          & 13.3         & 23.5          & \multicolumn{1}{c}{3.4}          \\ \cline{2-9} 
                        & March\cite{marcheggiani2016discrete}                      & 9.0          & 6.4          & 15.5 & 5.7          & 4.5          & 7.9          & \multicolumn{1}{c}{1.9}          \\ \cline{2-9} 
                         & UIE-March\cite{simon2019unsupervised}  & 19.5          & 13.3          & 36.7          & 30.6         & 24.1          & 42.1          & \multicolumn{1}{c}{11.5}          \\ \cline{2-9} 
                         & UIE-PCNN \cite{simon2019unsupervised}   & 19.7          & 14.0          & 33.4          & 26.6          & 20.8         & 36.8          & \multicolumn{1}{c}{9.4}          \\ \cline{2-9} 
                         & UIE-BERT  & 22.4          & 17.6          & 30.8          & 31.2         & 26.3          & 38.3          & \multicolumn{1}{c}{12.3}          \\ \cline{2-9} 
                        & {{\modelname} w/o Classification} & 31.5 & 23.2 & 49.1         & 14.1 & 10.9 & 19.8          & \multicolumn{1}{c}{7.7} \\ \cline{2-9}
                         & {{\modelname} w/o {\clustering}} & 32.0 & 26.3 & 41.0         & 16.9 & 14.3 & 20.8          & \multicolumn{1}{c}{12.7} \\ \cline{2-9}
                        & {\modelname}  & \textbf{32.9} & 29.7 & 36.8       & \textbf{32.4} & 30.1 & 35.1          & \multicolumn{1}{c}{\textbf{20.1}} \\
\end{tabular}
}
\caption{Quantitative performance evaluation on three datasets.}\label{tab:data}
\end{table*}
\subsection{Implementation Details}
Following the settings used in \citet{simon2019unsupervised}, all models are trained with 10 relation classes. Although it is lower than the number of true relations in the dataset, it still reveals important insights as the distribution of target relations is very unbalanced. Also, this allows us to do a fair comparison with baseline results.

For {\bert}, we use the default tokenizer in BERT to preprocess dataset and set max-length as 128. We use the pretrained \texttt{BERT-Base\_Cased} model to initialize parameters for {\bert} and use BertAdam to optimize the loss. 

For {\clustering}, we use an autoencoder with fully connected layers with the following dimensions ${2h_{R}}$-500-500-200 as the ${encoder}$ and 200-500-500-${2h_{R}}$ for the ${decoder}$. We randomly initialize weights using a Gaussian distribution with zero-mean and a standard deviation of 0.01. The autoencoder is pretrained for 20 epoches with $1e{-3}$ learning rate and $1e{-5}$ weight-decay with Adam Optimizer. To get the initial centroids, we applied $k$-means and set $K$ as 10.

For {\classification}, we use a fully connected layer as $c_{\tau}$ and set dropout rate to 10\%, learning rate to $1e{-5}$ and warm-up rate to 0.1. We fixed the parameters in $f_\theta$ for the first three epochs to allow the classification layer to warm up.

\subsection{Results}
Table \ref{tab:data} shows the experimental results. UIE-PCNN is considered as the previous state-of-the-art result. We enhance this baseline by replacing PCNN and GloVe embedding with the proposed BERT-based encoder and classifier. The enhanced state-of-the-art model, namely UIE-BERT, achieves the best performance among baselines.
The proposed {\modelname} model outperforms all baseline models consistently on B$^{3}$ F1/Precision, V-measure F1/Homogeneity and ARI. {\modelname} on average achieves 7.0\% higher in B$^{3}$ F1, 3.4\% higher in V-measure F1 and 7.7\% higher in ARI among three datasets when comparing with UIE-BERT. Unlike baseline methods which achieve high B$^{3}$ Recall but low Precision, or high V-measure Completeness but low Homogeneity, our model obtains a more balanced performance while achieving the highest Precision and Homogeneity, although B$^{3}$ Recall and V-measure Completeness are less satisfactory. Having high precision and homogeneity scores can be a quite appealing property for precision-oriented applications in the real-world.

\begin{figure*}[tbp!]
\centering
\includegraphics[width=0.26\textwidth]{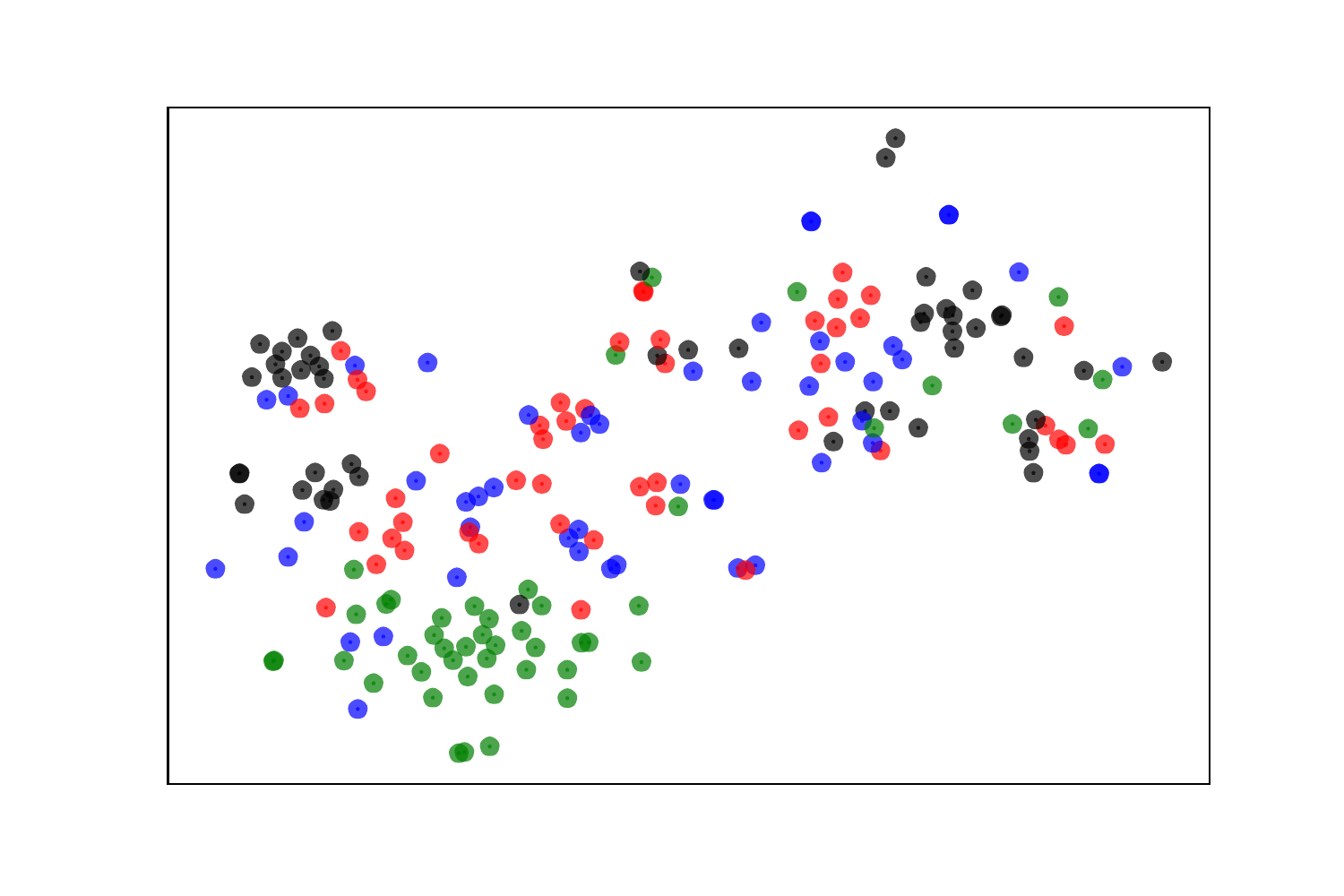}
\includegraphics[width=0.26\textwidth]{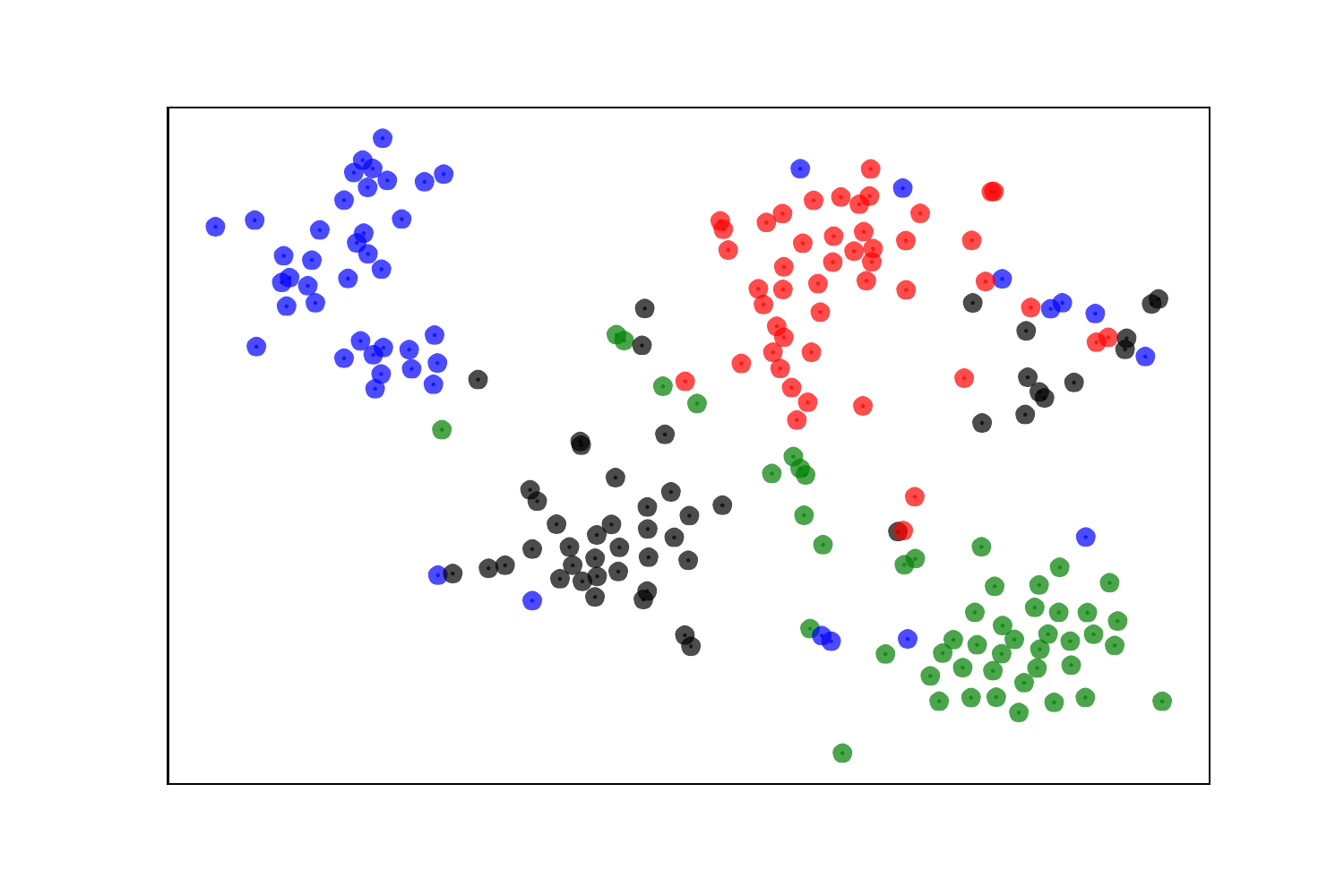}
\includegraphics[width=0.26\textwidth]{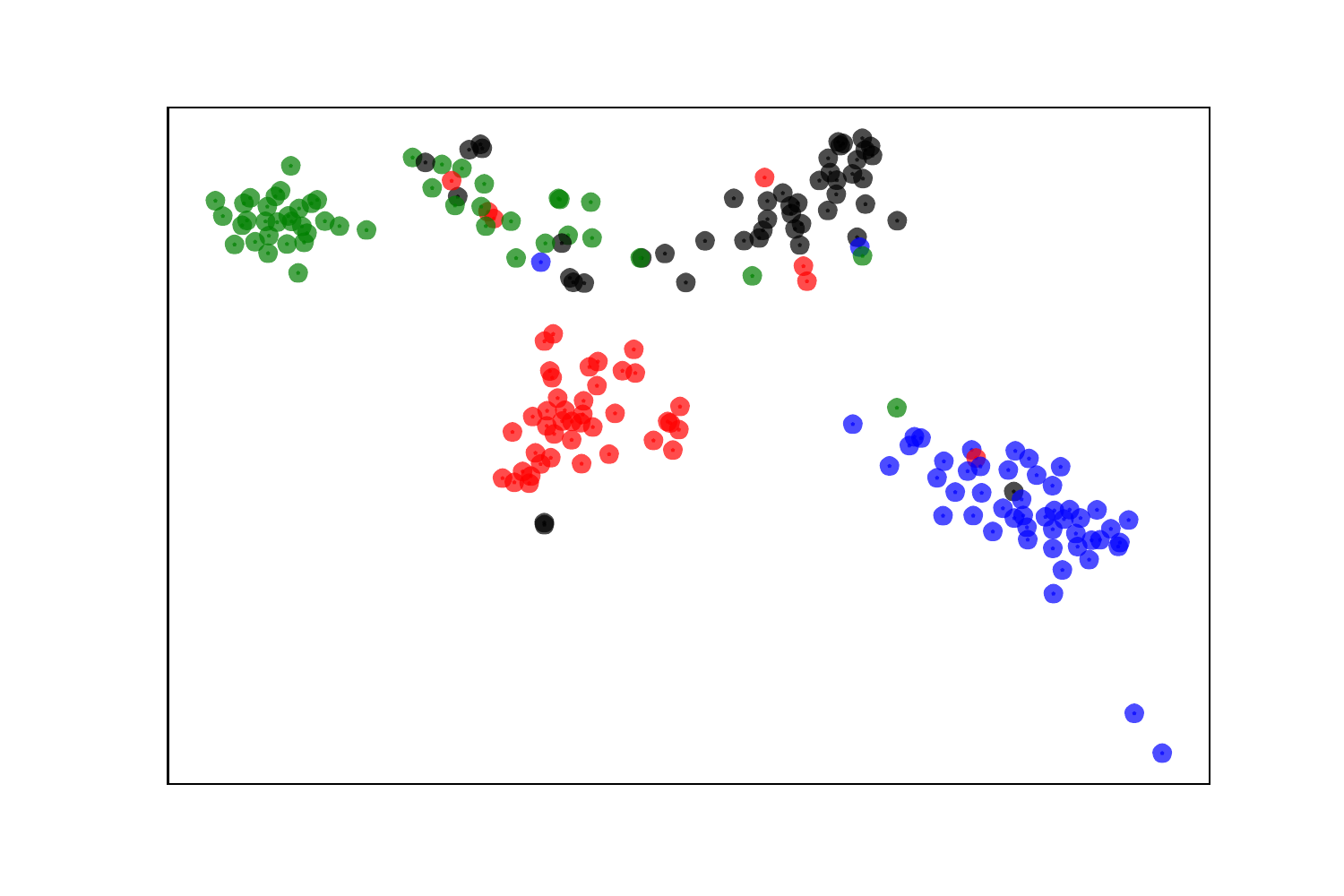}
\caption{Visualizing contextualized entity pair features after t-SNE dimension reduction for {\modelname} w/o classification (left), {\modelname} w/o {\clustering} (middle) and {\modelname} (right) on NYT+FB dataset.}\label{fig:tsne}
\end{figure*}

\noindent\textbf{Ablation Study}\\
We conduct ablation study to show the effectiveness of different modules of {\modelname} to the overall improved performance. {\modelname} w/o Classification is the proposed model without {\classification} and only uses the {\bert} for {\clustering}. {\modelname} w/o {\clustering} replaces the proposed soft-assignment clustering methods with $k$-means clustering as a hard-assignment alternative.

A general conclusion from ablation rows in Table \ref{tab:data} is that all modules contribute positively to the improved performance. More specifically, without self-supervised signals for relational feature learning, {\modelname} w/o Classification gives us 14.4\% less performance averaged over all metrics on all datasets. Similarly, {\clustering} gives 6.2\% performance boost in average over all metrics when comparing with the hard-assignment alternative ({\modelname} w/o {\clustering}).

\noindent\textbf{Visualize Contextualized Features}\\
To intuitively show how self-supervised learning helps learn better contextualized relational features on entity pairs, we visualize the contextual representation $\mathbb{R}^{2\cdot{h_{R}}}$ after dimension reduction using t-SNE \cite{maaten2008visualizing}. We randomly choose 4 relations from NYT+FB dataset and sample 50 entity pairs. The visualization results are shown in Figure \ref{fig:tsne}. Features are colored with their ground-truth relation labels.

From Figure \ref{fig:tsne} we can see that the features obtained through the raw BERT model (left) can already give meaningful semantics to entity pairs having different relations. But these features are not tailored for the relation extraction task. When {\clustering} is not applied (middle) and simply using $k$-means, which performs hard-assignment on samples, the proposed model without Adaptive Clustering gives decent results but does not provide confident cluster assignments. The proposed model (right) uses soft-assignment and a self-supervised learning schema to improve the relational feature learning ---we learn denser clusters and more discriminitaive features.

\vspace{0.1in}
\noindent\textbf{Sensitivity analysis: when K is unknown}\\
The {\clustering} gives {\modelname} enough flexibility to model relational features without knowing any prior information on the number of target relations or the relation distribution. This property is appealing when the number of target relations is not available for Relation Extraction on an open-domain corpus. 

The proposed model does require an intial cluster size $\hat K$ as the scope for pseudo labels. A general guideline for choosing $\hat K$ is to choose a value that is larger than the actual number of relations in the corpora as over-specifying the cluster size should not hurt the model performance. We set an initial ${\hat K}$ (for example ${\hat K}=1000$), and use an unsupervised method, here we use $k$-means, to merge $\hat K$ cluster centroids into $K$ clusters for evaluation. 

\begin{figure}[hbp!]
    \centering
    \includegraphics[width=0.85\linewidth]{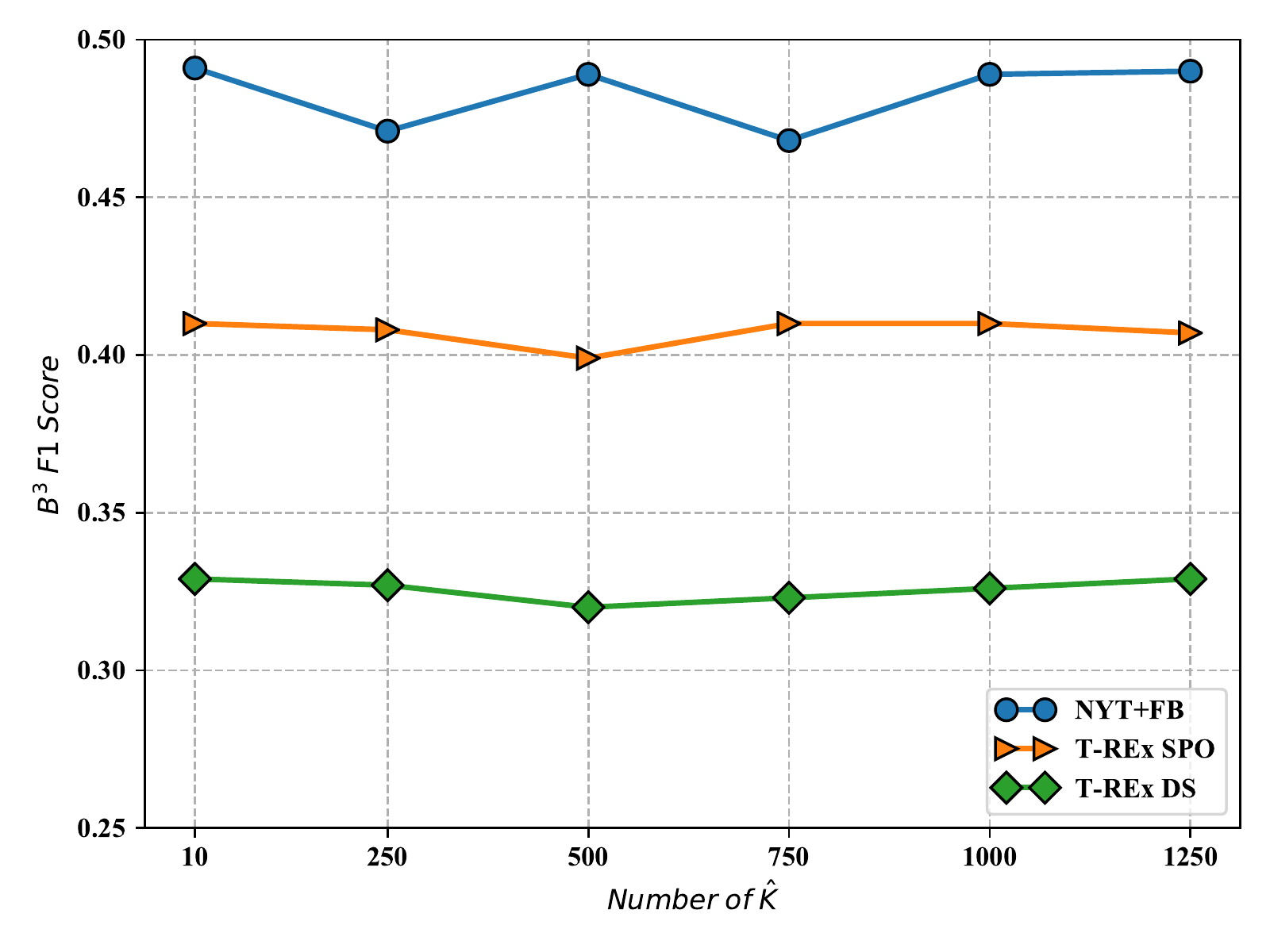}
    \vspace{-0.05in}
    \caption{F1 Score with different $\hat K$.}
    \label{fig:stable}
\end{figure}

We vary $\hat K$ from 10 to 1250 and report the B$^{3}$ F1 score when comparing the predicted relation type (based on $K$ clusters after merging) with the golden relation type. As shown in Figure \ref{fig:stable}, the best performance is obtained when $\hat K=10$, indicating that {\modelname} actually leverages the number of target relations as a useful prior knowledge. Thanks to the self-learning schema and the {\clustering}, when we very $\hat K$ from 10 to 1250, the model achieves stable F1 score and is not sensitive to the initial choice of $\hat K$ on all three datasets. The results also further indicate the applicability of the proposed model when being applied to an open-domain corpus when the number of target relations is not available in advance. We can assign a larger $\hat K$ value than needed and the model is still robust. Note that merging $\hat K$ clusters into $K$ clusters is mainly for evaluation purpose: when $K$ is unknown in advance and we simply use a large ${K}$ directly, it does result in $K$ clusters where clusters tend to be smaller, and multiple clusters may correspond to entity pairs having the same relation. 

\noindent\textbf{Surface-form Relation Names}\\
We provide a brief case study to show the surface-form relation names we extracted for each cluster (introduced in Section \ref{sec:rel_name}). We randomly select 5 relations in T-REx SPO and report the extracted surface-form relation names using frequent n-gram in Table \ref{tab:rel_name}.
\begin{table}[tbp!]
    \centering
    \resizebox{0.95\linewidth}{!}{%
    \begin{tabular}{c|c}
        Extracted surface-form & Golden surface-form \\\hline
        are close to & shares border with\\
        the state of & country\\
        capital city & capital\\
        son of & child\\
        member of & member of\\
    \end{tabular}
    }\vspace{-0.05in}
    \caption{Extracted and golden surface-form relation names on T-REx SPO.}
    \label{tab:rel_name}
    \vspace{-0.1in}
\end{table}
The surface-form relation name extraction gives {\modelname} an extended ability to not only discriminate between entity pairs having different relations, but also derive surface-forms for relation clusters as the final Relation Extraction results. However, evaluating the quality of relation surface-forms is out-of-scope for this work.

\vspace{-0.05in}
\section{Related Works}
\vspace{-0.05in}
Relation extraction focuses on identifying the relation between two entities in a given sentence. Traditional closed-domain relation extraction methods are supervised models. They need a set of pre-defined relation labels and require large amounts of annotated triplets, making them less ideal 
to work on open-domain corpora.
Distant supervision \cite{mintz2009distant,surdeanu2012multi} is a widely used method to alleviate human annotation: if multiple sentences contain two entities that have a certain relation in a knowledge base, at least one sentence is believed to convey the related relation. However, entities convey semantic meanings also according to the contexts, distant supervised models do not explicitly consider contexts and the models cannot discover new relations as the supervision is purely adopted from knowledge bases.

Unsupervised relation extraction \cite{stanovsky2018supervised,saha2018open,yu2017open} gets lots of attention, due to the ability to discover relational knowledge without access to annotations and external resources.
Unsupervised models either 1) cluster the relation representation extracted from the sentence; 2) make more assumptions that provide learning signals for classification models. Among clustering models, an important milestone is the OpenIE approach \cite{banko2007open}, assuming the surface form of relations will appear between two entities in its dependency tree. However, these works heavily rely on surface-form relation and have less ideal generalization capabilities. To solve this problem, \citet{roy2019supervising} proposes a system that learns to supervise unsupervised OpenIE model, which combines the strength and avoids the weakness in each individual OpenIE system. Relation knowledge transfer system \cite{wu2019open} learns similarity metrics of relations from labeled data, and then transfers the relational knowledge to identify novel relations in unlabeled data. 

\citet{marcheggiani2016discrete} proposes a variational autoencoder approach (VAE): the encoder part extracts relations from labeled features, and the decoder part predicts one entity when given the other entity and the relation with the function of triplet scoring \cite{nickel2011three}. This scoring function could provide a signal since it is known to predict relation triplets when given their embeddings. However, posterior distribution and prior uniform distribution based on KL divergence is unstable. \citet{simon2019unsupervised} proposes a model to solve instability and trains the features on classifiers such as PCNN model \cite{zeng2015distant}.

Inspired by the success of self-supervised learning in computer vision \cite{wiles2018self,caron2018deep}, and large pretrained language models that show great potential in encoding meaningful semantics for various downstream tasks \cite{devlin2018bert}, we proposed a self-supervised learning schema for open-domain relation extraction. It has the advantages of unsupervised learning to handle the cases where the number of relations is not known in advance, but also keeps the advantage of supervised learning that has strong discriminative power for relational feature learning.
\vspace{-0.05in}
\section{Conclusions}
\vspace{-0.1in}
We propose a self-supervised learning model {\modelname} for open-domain relation extraction.
Different from conventional distant-supervised models which require labeled instances for Relation Extraction in a closed-world setting, our model does not require annotations and is able to work on open-domain scenarios when target relation number and relation distributions are not known in advance.
Comparing with unsupervised models, our model exploits the advantages of supervised models and bootstraps the discriminative power using self-supervised signals via learning improved contextualized relational features.
Experiments on three real-world datasets show effectiveness and robustness of {\modelname} over competitive baselines.
\section*{Acknowledgments}
We thank the reviewers for their valuable comments. We thank Diego Marcheggiani for sharing the NYT+FB dataset. The work was supported by the National Key Research and Development Program of China (No.2019YFB1704003), the National Nature Science Foundation of China (No. 71690231), NSF under grants III-1763325, III-1909323, SaTC-1930941 and Tsinghua BNRist.

\balance
\bibliography{anthology,emnlp2020}
\bibliographystyle{acl_natbib}

\appendix
\end{document}